\ifdefined\XeTeXversion\else
  \ifdefined\pdfoutput
    \pdfoutput=1
  \fi
\fi
\documentclass[10pt,twocolumn,a4paper]{article}

\usepackage[margin=0.72in,top=0.68in,bottom=0.72in,columnsep=0.26in]{geometry}
\usepackage[authoryear,round]{natbib}
\usepackage{amsmath,amssymb}
\usepackage{graphicx}
\usepackage{booktabs}
\usepackage{tabularx}
\usepackage{array}
\usepackage{makecell}
\usepackage{adjustbox}
\usepackage{enumerate}
\usepackage{placeins}
\usepackage{xcolor}
\usepackage{balance}
\usepackage{microtype}
\usepackage[colorlinks=true,allcolors=blue!55!black]{hyperref}

\hypersetup{
  pdftitle={EGM-Det: Entropy-Guided Multimodal Adaptive Fusion for UAV RGB-IR Object Detection},
  pdfauthor={Cunzheng Fan, Dawei Yan, Guanlin Wang, Xingshuo Yang, Yupeng Jia, Jing Yang, Haokui Zhang}
}
\newcommand{\tabsetup}{%
  \small
  \setlength{\tabcolsep}{4pt}
  \renewcommand{\arraystretch}{1.12}
}

\newcommand{\legendbox}[1]{\textcolor{#1}{\rule{1.2em}{0.7em}}}


\begin{document}

\twocolumn[
\begin{@twocolumnfalse}
\begin{center}
  {\LARGE\bfseries EGM-Det: Entropy-Guided Multimodal Adaptive Fusion for UAV RGB--IR Object Detection\par}
  \vspace{0.8em}
  {\large
  Cunzheng Fan$^{1,\dagger}$,
  Dawei Yan$^{1,\dagger}$,
  Guanlin Wang$^{1}$,
  Xingshuo Yang$^{1}$,
  Yupeng Jia$^{1}$,
  Jing Yang$^{2,*}$, and
  Haokui Zhang$^{1,*}$\par}
  \vspace{0.55em}
  {\small
  $^1$School of Cybersecurity, Northwestern Polytechnical University, Xi'an 710072, China\\
  $^2$School of Automation and Software Engineering, Shanxi University, Taiyuan 030006, China\par}
  \vspace{0.35em}
  {\footnotesize
  $^\dagger$These authors contributed equally to this work.\\
  $^*$Corresponding authors:
  \href{mailto:yangjing199002@sxu.edu.cn}{\texttt{yangjing199002@sxu.edu.cn}};
  \href{mailto:hkzhang@nwpu.edu.cn}{\texttt{hkzhang@nwpu.edu.cn}}\par}
\end{center}

\begin{abstract}
Joint utilization of RGB and infrared (IR) information for object detection represents an important research direction in UAV-view object detection. Multimodal complementarity alleviates indistinct UAV-view target features and greatly improves detection accuracy. However, most existing methods simply fuse multimodal features via static or fixed spatial weights, ignoring the varying contributions of each modality across diverse scenarios and failing to fully exploit multimodal complementarity.
In this paper, we propose EGM-Det, an entropy-guided multimodal adaptive fusion framework for RGB-IR object detection. Guided by information entropy, the proposed EGM-Det designs scenario-adaptive fusion strategies to fully exploit multimodal complementarity and boost detection performance.
Specifically, EGM-Det adopts a dual-stream architecture to preserve modality-specific representations and introduces an Entropy Offset Gate Fusion module for adaptive multi-scale fusion. This module constructs shallow entropy priors from input intensity, local entropy, and cross-modal discrepancy, and uses them to guide local offset alignment and spatial-channel gated fusion. In this way, EGM-Det selectively aggregates reliable RGB and infrared cues rather than uniformly combining heterogeneous features.
Furthermore, cross-modal distillation is proposed to regularize learned fusion gates for avoiding fusion degradation. Distinct from prior distillation methods, our mechanism lets each student branch extract complementary knowledge from its cross-modality teacher branch matched to the main branch, further amplifying the advantages of multimodal fusion.
We conduct experiments on three representative benchmarks: DroneVehicle, LLVIP and VEDAI. Experimental results demonstrate that our proposed EGM-Det achieves state-of-the-art performance across all three datasets. In particular, our method surpasses prior approaches by over 10 percentage points on the VEDAI benchmark.
\end{abstract}

\vspace{1.0em}
\end{@twocolumnfalse}
]

\section{Introduction}
UAV-view object detection has become an important perception task for traffic monitoring, urban surveillance, and emergency response~\citep{zhu2018vision,sun2020drone,cheng2016survey}. Compared with ground-view images, aerial images usually contain small, densely distributed, and arbitrarily oriented objects under complex backgrounds and variable illumination~\citep{xia2018dota,ding2021object,ding2019learning,yang2021r3det}. These factors make robust vehicle detection difficult, especially when object appearance changes with viewpoint, altitude, illumination, and imaging quality. RGB-infrared (RGB-IR) multimodal imaging provides a promising solution by combining the texture and structural details of visible images with the illumination robustness and thermal responses of infrared images~\citep{sun2020drone,jia2021llvip}. Recent UAV-oriented infrared-visible fusion studies further show that cross-modal fusion is critical for robust perception under low illumination, backlighting, haze, smoke, scale variation, and onboard resource constraints~\citep{li2025infrared}. However, effectively exploiting such complementary information remains non-trivial because the reliability of RGB and IR cues can vary substantially across spatial locations and imaging conditions.

The central challenge of RGB-IR detection lies in the spatially varying reliability of different modalities. This challenge arises from two coupled factors. First, RGB and IR images are generated by different imaging mechanisms, leading to heterogeneous texture, contrast, and thermal responses~\citep{sun2020drone,fang2021crossmodality}. Second, modality reliability is inherently content-dependent: a region that is discriminative in RGB may be ambiguous in IR, whereas another region may benefit more from thermal cues~\citep{li2019illumination,fang2021crossmodality}.
A reliable detector should learn not only how to combine cross-modal features, but also when and where each modality should be trusted.

Existing RGB-IR and visible-thermal detectors have explored
attention-based feature fusion~\citep{fang2021crossmodality,shen2024icafusion,wu2026lightweight}, Transformer-based cross-modal interaction~\citep{fang2021cross,yuan2024c2former}, end-to-end synchronous fusion and detection~\citep{zhang2024e2emfd}, and conflict-aware multimodal learning~\citep{he2023multispectral}. Although these methods improve multimodal representation learning, the modality-selection process is generally optimized only through the final detection objective. Such implicit supervision does not directly indicate which modality is more reliable at a specific spatial location or feature level. Consequently, unreliable or redundant modality responses may still be propagated when RGB and IR reliability varies across locations and imaging conditions. This motivates explicit modeling and supervision of modality preference for robust RGB-IR object detection.

Knowledge distillation provides a possible way to guide multimodal student learning with teacher models~\citep{hinton2015distilling}. Conventional detection distillation usually transfers feature-level or prediction-level knowledge from teacher detectors to a student detector~\citep{dai2021general,yang2021focal,wang2023crosskd}. However, RGB-IR multimodal detection requires more than mimicking final predictions or fused features. The student must preserve modality-specific representations, exploit cross-modal complementarity, and learn adaptive modality preference simultaneously. In this setting, the key question is not only how to distill detection knowledge, but also how to distill reliable modality selection behavior. This motivates a selective distillation strategy that focuses supervision on the components most relevant to multimodal fusion.

To address these problems, we propose EGM-Det, an entropy-guided multimodal adaptive fusion framework for RGB-infrared object detection. Centered on information entropy, EGM-Det leverages scenario-adaptive fusion to exploit multimodal complementary information. It adopts a dual-stream architecture and introduces an Entropy Offset Gate Fusion (EOGF) module for adaptive multi-scale fusion. This module builds shallow entropy priors from intensity values, local entropy and cross-modal discrepancies to guide local offset alignment and spatial-channel gated aggregation, selectively integrating reliable RGB and infrared features instead of static uniform fusion. To mitigate fusion degradation, we further propose a cross-modal distillation strategy to regularize fusion gates. Unlike existing distillation schemes, each student branch extracts complementary knowledge from its matched cross-modal teacher branch, which further strengthens the benefits of multimodal fusion.

The main contributions of this paper are summarized as follows:
\begin{enumerate}[(1)]
    \item We propose EGM-Det, where RGB-IR information is integrated by an entropy-guided alignment-aware gated fusion module. It achieves more flexible multimodal fusion, dynamically perceiving scene properties to determine the dominant modality and further exploit the complementary strengths across modalities.

    \item We introduce a dual-teacher modality-preference distillation strategy. Relative dense confidence responses from RGB and IR teachers are converted into soft modality-preference targets, and student gate entropy is further used to adaptively strengthen gate supervision in ambiguous regions.
    \item We refine the DroneVehicle annotations by correcting cross-modal inconsistencies and category-level errors, establish a refined paired RGB-IR evaluation protocol, and validate EGM-Det on the refined DroneVehicle and other two representive datasets LLVIP and VEDAI. Our proposed EGM-Det achieve the best performance on all three datasets.
\end{enumerate}
\section{Related Work}
\subsection{UAV-View RGB-IR and Visible-Thermal Object Detection}
UAV-view object detection has been widely studied for traffic monitoring, urban surveillance, and emergency perception. Compared with ground-view detection, UAV-view detection is more challenging because objects are often small, densely distributed, arbitrarily oriented, and captured under complex backgrounds \citep{yang2021r3det, ding2019learning}. To improve robustness under illumination variation and degraded visible appearance, RGB-infrared and RGB-thermal detection have attracted increasing attention. Visible images provide texture, color, and structural details, while infrared or thermal images are less sensitive to illumination changes and can highlight objects with thermal responses.

Several datasets have supported this research direction~\citep{hwang2015multispectral}. DroneVehicle is a representative drone-based RGB-infrared vehicle detection benchmark, which provides paired RGB-IR images and oriented vehicle annotations for cross-modality detection~\citep{sun2020drone}. VEDAI focuses on vehicle detection in aerial imagery and is commonly used for evaluating small vehicle detection under unconstrained aerial conditions~\citep{razakarivony2016vehicle}. LLVIP provides aligned visible-infrared image pairs for low-light vision, and has been used in visible-infrared fusion and detection studies~\citep{jia2021llvip}. Large-scale aerial detection benchmarks such as DOTA, VisDrone and DIOR further highlight the challenges of small objects, dense layouts, and arbitrary object orientations in aerial imagery~\citep{xia2018dota,ding2021object,zhu2018vision,li2020object}. These datasets show that multimodal and aerial object detection are valuable under challenging imaging conditions, but they also reveal two practical issues: cross-modal observations may contain redundant or conflicting responses, and paired annotation consistency is critical for reliable multimodal training and evaluation.

Existing RGB-IR and RGB-T detectors mainly focus on improving
multimodal feature fusion. Recent methods have explored
attention-based feature fusion
~\citep{fang2021crossmodality,shen2024icafusion,wu2026lightweight},
Transformer-based cross-modal interaction
~\citep{fang2021cross,yuan2024c2former},
end-to-end synchronous fusion and detection
~\citep{zhang2024e2emfd},
and conflict-aware multimodal learning
~\citep{he2023multispectral}.
Other studies investigate mono-modality feature learning
~\citep{zhao2025rethinking}
and alignment-oriented or offset-guided feature modeling
~\citep{yuan2022translation,liu2026cross,liu2025cross}
to alleviate modality gaps and local feature inconsistency. For example, UA-CMDet models uncertainty in DroneVehicle-based cross-modality vehicle detection~\citep{sun2020drone}, and TSFADet addresses weak cross-modal misalignment in aerial RGB-IR vehicle detection through translation-, scale-, and rotation-aware alignment~\citep{yuan2022translation}. Recent benchmark-oriented studies also investigate visible-thermal misalignment and provide dedicated baselines for drone-based visible-thermal object detection~\citep{song2024misaligned}. More recent UAV-oriented methods further explore offset-guided fusion and dynamic alignment for weakly aligned multimodal detection~\citep{liu2026cross,liu2025cross}. These studies show that RGB-IR detection requires more than direct feature fusion, especially when the two modalities are weakly aligned, locally inconsistent, or unevenly reliable.

However, most existing methods still learn modality fusion mainly through detection supervision. Although such supervision improves final detection accuracy, it does not explicitly indicate which modality is more reliable at a specific spatial location or feature level. As a result, the fusion module may still propagate unreliable modality responses when RGB and IR cues are locally inconsistent. This limitation is particularly important for UAV-view RGB-IR detection, where modality reliability can change sharply with illumination, background clutter, target scale, and local imaging quality. In contrast, EGM-Det treats modality selection as an explicit distillation target. It derives modality-preference distributions from dual-teacher confidence maps and further uses student gate entropy to strengthen supervision for uncertain regions.

\begin{figure*}[t]
    \centering
    \includegraphics[width=0.8\textwidth]{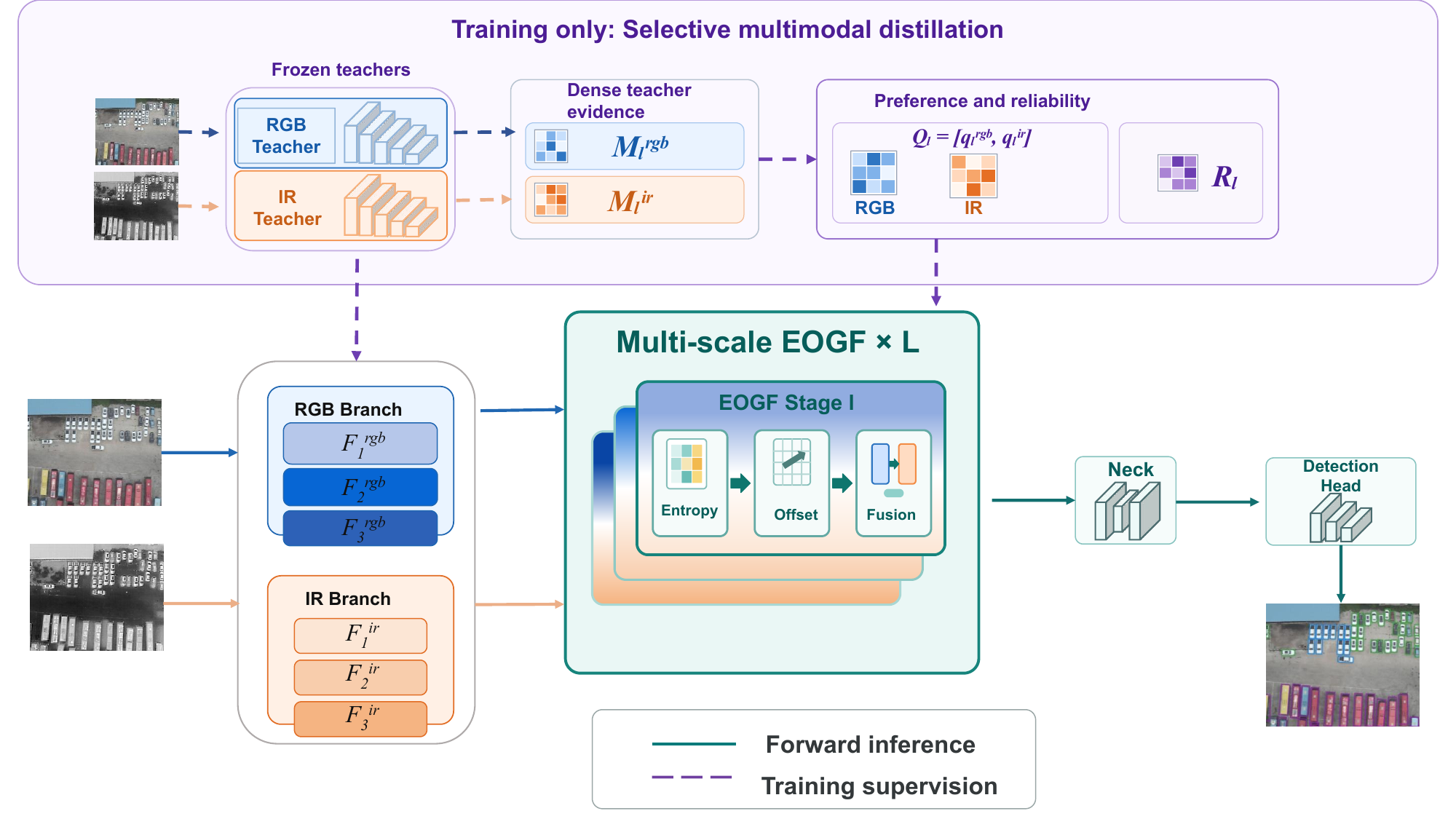}
    \vspace{-2mm}
    \caption{
Overview of EGM-Det. The dual-stream detector extracts modality-specific RGB and infrared features and applies multi-scale EOGF modules for entropy-guided alignment and gated fusion. During training, two frozen unimodal teachers provide dense confidence maps, from which modality-preference targets and reliability weights are constructed to supervise the spatial gates and feature branches. Only the dual-stream detector is retained during inference.
}
    \label{fig:over_view}
\end{figure*}

\subsection{Knowledge Distillation for Object Detection}

Knowledge distillation transfers knowledge from a high-capacity teacher model to a compact or task-specific student model~\citep{hinton2015distilling}. In object detection, KD is more complex than in image classification because a detector must jointly handle classification, localization, multi-scale representation, and foreground-background imbalance. Existing detection KD methods have therefore explored different forms of knowledge, including channel-wise dense prediction distillation, feature imitation, instance-level distillation, relation distillation, and prediction mimicking~\citep{shu2020channel,guo2021distilling,dai2021general,yang2021focal,wang2023crosskd,zagoruyko2017paying,chen2017learning}.

Representative methods show that detection distillation should be selective and task-aware. GID selects discriminative instances and combines feature-, relation-, and response-based knowledge for detector distillation~\citep{dai2021general}. FGD observes that teacher-student feature differences vary across foreground/background regions and proposes focal and global distillation to emphasize important pixels, channels, and global relations~\citep{yang2021focal}. CrossKD revisits prediction mimicking by passing student head features through the teacher head, reducing contradictory supervision between annotations and teacher predictions~\citep{wang2023crosskd}. These methods demonstrate that effective detection KD depends on where and what to distill, rather than applying uniform imitation to all features or predictions~\citep{wang2019distilling}.

Although these methods are effective for general object detection, they are mainly designed for single-modality detectors. In RGB-IR multimodal detection, the student must learn modality-specific representations, cross-modal complementarity, and adaptive modality preference. Directly distilling fused features or detection heads may be suboptimal, because the multimodal student should not simply mimic a teacher's final response. Instead, it should learn how to select reliable cues from different modalities. EGM-Det follows this principle by adopting a selective distillation objective that preserves branch-level, cross-modal, and gate-level supervision while excluding ineffective pseudo-fusion and detection-head distillation branches. The proposed student-gate-entropy weighting further makes this process uncertainty-aware, allowing stronger supervision to be assigned to ambiguous modality decisions.
\section{Method}

\subsection{Overview of EGM-Det}

EGM-Det models spatially varying modality reliability rather than assuming uniformly reliable RGB and infrared features. As illustrated in Figure~\ref{fig:over_view}, paired RGB--IR images are processed by two modality-specific branches to extract multi-scale representations. At selected feature levels, the proposed EntropyOffsetGateFusion (EOGF) module uses shallow priors derived from image intensity, local entropy, and cross-modal discrepancy to guide local feature alignment and spatial and channel gating. During training, selective multimodal distillation provides auxiliary supervision for the spatial modality gates, whereas only the dual-stream detector is retained for inference.

\subsection{Entropy-Guided Dual-Stream Architecture}
\label{sec:dual-stream}
\begin{figure*}[t]
    \centering
    \includegraphics[width=0.95\textwidth]{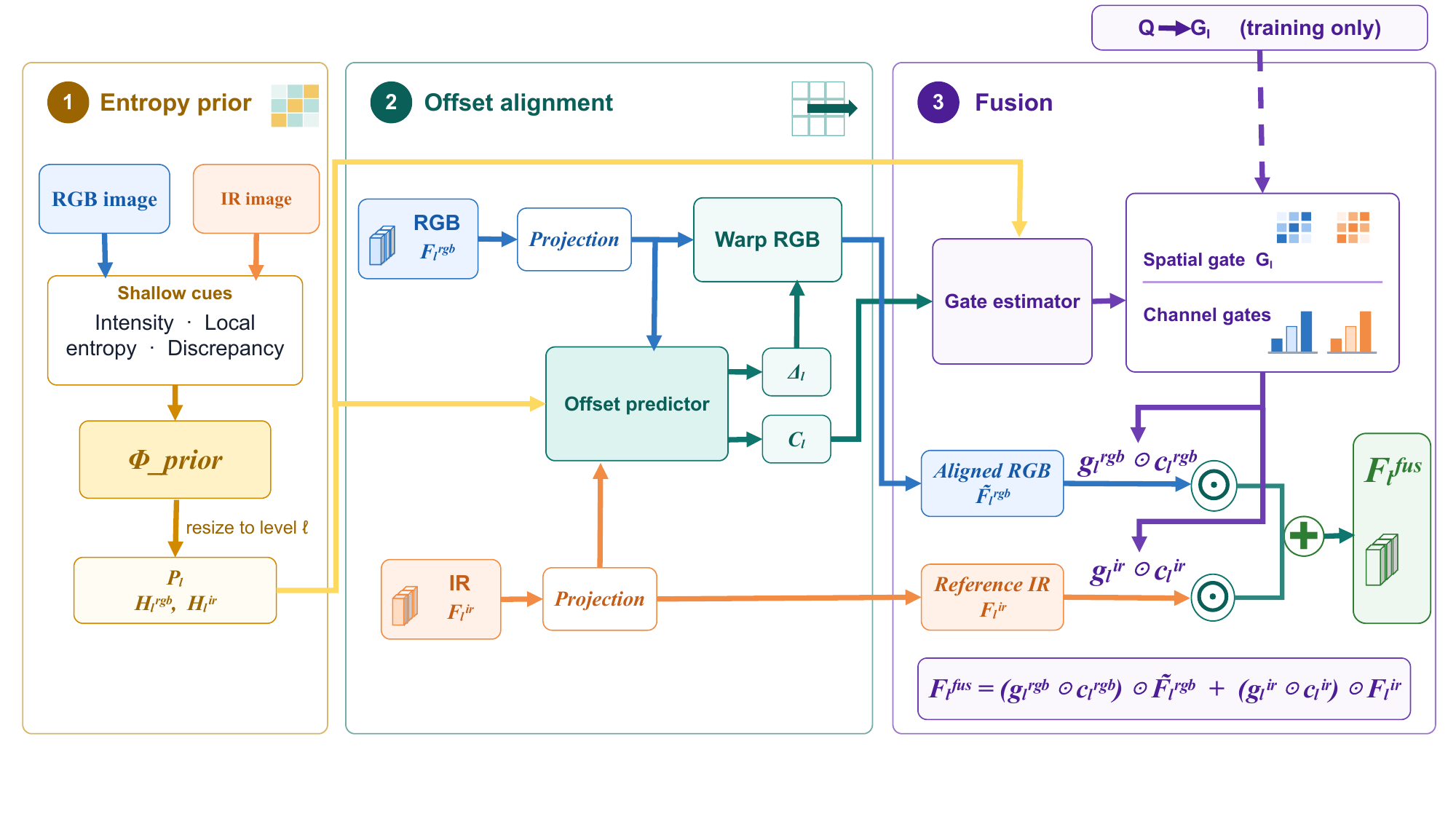}
    \vspace{-2mm}
\caption{
Detailed structure of EntropyOffsetGateFusion at feature level \(l\). The module first constructs level-specific entropy cues \(P_l\), \(H_l^{rgb}\), and \(H_l^{ir}\) from input intensity, local entropy, and cross-modal discrepancy. Guided by \(P_l\), the projected features are used to estimate an offset field \(\Delta_l\) and alignment confidence \(C_l\), after which the RGB feature is aligned to the IR reference. The gate estimator then predicts the spatial gate \(G_l\) and channel gates \(c_l^{rgb}\) and \(c_l^{ir}\) to produce the fused feature \(F_l^{fus}\). The dashed purple arrow denotes training-only supervision of \(G_l\) by \(Q_l\).
}
\label{fig:entropygate}
\end{figure*}

RGB and infrared images provide complementary appearance and thermal cues, but their relative reliability varies across spatial locations. Direct early fusion may mix unreliable responses before modality-specific representations are sufficiently formed. As illustrated in Figure~\ref{fig:entropygate}, EGM-Det therefore adopts two symmetric modality-specific branches to extract multi-scale RGB and infrared features, denoted by \(F_l^{rgb}\) and \(F_l^{ir}\), respectively. Cross-modal interaction is introduced only at selected feature levels through the proposed EntropyOffsetGateFusion (EOGF) module. This design preserves modality-specific representations while allowing complementary information to be exchanged at multiple scales.

EOGF first constructs an input-level entropy prior to provide scene-dependent guidance for feature fusion. Let \(I^{rgb}\) and \(I^{ir}\) denote the normalized intensity maps, and let \(H^{rgb}\) and \(H^{ir}\) denote their local entropy maps. The entropy prior is defined as
\begin{equation}
P =
\Phi_{\mathrm{prior}}
\left(
I^{rgb}, I^{ir},
H^{rgb}, H^{ir},
\left|I^{rgb}-I^{ir}\right|
\right),
\end{equation}
where \(\Phi_{\mathrm{prior}}(\cdot)\) is a lightweight prior estimator. The resulting prior encodes local structural complexity, modality saliency, and cross-modal discrepancy. It is resized to each fusion level as \(P_l\) and used only to condition subsequent alignment and gating, rather than serving as a predefined modality-selection mask.

At feature level \(l\), \(F_l^{rgb}\) and \(F_l^{ir}\) are projected into a shared embedding space as \(Z_l^{rgb}\) and \(Z_l^{ir}\). To compensate for local spatial inconsistencies between the two modalities, EOGF predicts an offset field \(\Delta_l\) together with an alignment confidence \(C_l\):
\begin{equation}
\begin{aligned}
(\Delta_l,C_l)
&=
\Phi_{\mathrm{align}}
\left(
Z_l^{rgb},Z_l^{ir},
\left|Z_l^{rgb}-Z_l^{ir}\right|,
P_l
\right), \\
\widetilde{Z}_l^{rgb}
&=
\mathcal{W}(Z_l^{rgb},\Delta_l), \qquad
\widetilde{F}_l^{rgb}
=
\mathcal{W}(F_l^{rgb},\Delta_l),
\end{aligned}
\end{equation}
where \(\mathcal{W}(\cdot,\cdot)\) denotes differentiable warping~\citep{jaderberg2015spatial,dai2017deformable}. Unlike global image registration, this operation performs local correction in the feature space. The alignment confidence is subsequently provided to the gate estimator so that modality selection can account for the reliability of the estimated correspondence.

After alignment, EOGF predicts a two-way spatial modality gate:
\begin{equation}
\begin{aligned}
G_l &= [g_l^{rgb},g_l^{ir}],\\
G_l &= \operatorname{Softmax}\!\left[
\Phi_{\mathrm{gate}}\!\left(
\widetilde{Z}_l^{rgb},Z_l^{ir},H_l^{rgb},H_l^{ir},P_l,C_l
\right)\right].
\end{aligned}
\end{equation}
where \(g_l^{rgb}+g_l^{ir}=1\) at each spatial location. In parallel, channel gates \(c_l^{rgb}\) and \(c_l^{ir}\) estimate the importance of individual feature channels~\citep{hu2018squeeze,woo2018cbam}. The fused feature is obtained as
\begin{equation}
F_l^{fus}
=
g_l^{rgb}\odot c_l^{rgb}\odot\widetilde{F}_l^{rgb}
+
g_l^{ir}\odot c_l^{ir}\odot F_l^{ir},
\end{equation}
where \(\odot\) denotes element-wise multiplication. EOGF thus combines entropy-guided reliability cues, local feature alignment, and spatial-channel selection within a unified fusion module. The resulting multi-scale fused features are forwarded to the subsequent detection network, while \(G_l\) provides an explicit modality-preference distribution that is further regularized by the selective distillation strategy described in Section~\ref{sec:selective_distillation}.

\subsection{Selective Multimodal Distillation}
\label{sec:selective_distillation}
\begin{figure*}[t]
    \centering
    \includegraphics[width=0.95\textwidth]{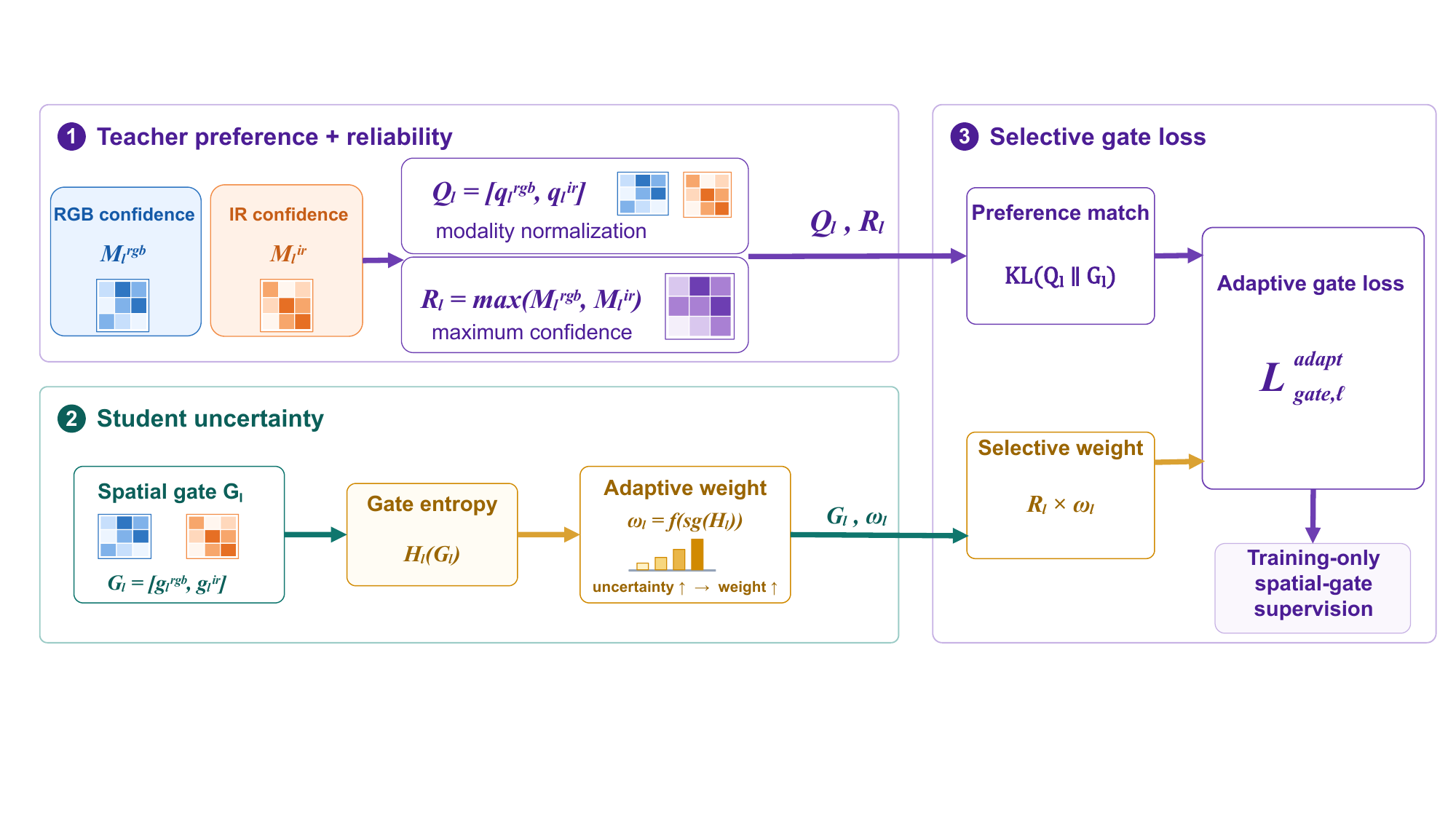}
    \vspace{-2mm}
    \caption{
Dual-teacher modality-preference construction and entropy-adaptive spatial-gate distillation. The RGB and IR teacher confidence maps are converted into a soft modality-preference target \(Q_l\), while their maximum response defines the reliability weight \(R_l\). The entropy \(H_l\) of the student spatial gate \(G_l\) is further converted into the pixel-wise adaptive weight \(\omega_l\). The preference-matching loss \(\mathrm{KL}(Q_l\parallel G_l)\) is weighted by \(R_l\omega_l\) to obtain \(\mathcal{L}_{gate}^{adapt}\). This supervision is applied only to the spatial gate during training, while the channel gates remain optimized by the detection and feature-distillation objectives.
}
\label{fig:teacher_gate_distillation}
\end{figure*}

During distillation, the entropy-guided dual-stream detector described in
Section~\ref{sec:dual-stream} serves as the student, while two independently
trained and frozen unimodal detectors serve as the RGB and infrared teachers.
Rather than transferring all teacher predictions, EGM-Det uses their relative
dense confidence responses to supervise the student's spatial modality gates.
As illustrated in Figure~\ref{fig:teacher_gate_distillation}, the supervision
is applied selectively: low-confidence teacher evidence is down-weighted,
whereas uncertain student decisions receive stronger guidance.

Let \(\mathcal{M}=\{rgb,ir\}\) denote the two modalities. At each feature level corresponding to a spatial gate, a dense confidence map is computed from the classification output of teacher \(m\in\mathcal{M}\):
\begin{equation}
M_l^m(x,y)
=
\max_{c\in\mathcal{C}}
\sigma\left(s_{l,c}^m(x,y)\right),
\label{eq:teacher_confidence}
\end{equation}
where \(s_{l,c}^m(x,y)\) denotes the class logit for category \(c\). Because
the adopted detection head has no separate objectness branch, the maximum
class confidence is used as the dense teacher response. The confidence map
is resized to the corresponding gate resolution and denoted by
\(\bar{M}_l^m\).

The relative teacher responses are then normalized into a soft modality-
preference distribution:
\begin{equation}
q_l^m(x,y)
=
\frac{
\max\left(\bar{M}_l^m(x,y),\epsilon\right)
}{
\displaystyle
\sum_{n\in\mathcal{M}}
\max\left(\bar{M}_l^n(x,y),\epsilon\right)
},
\qquad m\in\mathcal{M},
\label{eq:teacher_preference}
\end{equation}
where \(Q_l=[q_l^{rgb},q_l^{ir}]\) is the teacher-derived modality-preference
target. It represents the relative reliability of RGB and infrared cues at
each spatial location. The modality-preference target \(Q_l\) is constructed directly from the dense classification responses of the two teachers, avoiding additional processing of predicted bounding boxes. It supervises only the spatial gate \(G_l\), while the channel
gates remain optimized by the detection and feature-distillation objectives.

A fixed gate-distillation weight cannot distinguish an unreliable teacher
target from a genuinely ambiguous student decision. We therefore quantify
these two factors separately. Given the student gate
\(G_l=[g_l^{rgb},g_l^{ir}]\), its modality-selection uncertainty is measured
by the normalized entropy
\begin{equation}
H_l(x,y)
=
-\frac{1}{\log 2}
\sum_{m\in\mathcal{M}}
g_l^m(x,y)
\log\left(g_l^m(x,y)+\epsilon\right).
\label{eq:gate_entropy}
\end{equation}
A high value of \(H_l\) indicates that the student has not formed a clear
preference between the two modalities.

The reliability of the teacher target is measured as
\begin{equation}
R_l(x,y)
=
\max\left(
\bar{M}_l^{rgb}(x,y),
\bar{M}_l^{ir}(x,y)
\right).
\label{eq:teacher_reliability}
\end{equation}
Thus, locations at which both teachers produce weak responses contribute less
to gate distillation.

The student uncertainty is further converted into a pixel-wise adaptive weight:
\begin{equation}
\resizebox{0.96\columnwidth}{!}{$\displaystyle
\begin{aligned}
\omega_l(x,y)
=
\operatorname{clip}\Big(
&1+
\beta_{\mathrm{pos}}
\left[
\operatorname{sg}\!\left(H_l(x,y)\right)-\tau
\right]_{+} \\
&-
\beta_{\mathrm{neg}}
\left[
\tau-\operatorname{sg}\!\left(H_l(x,y)\right)
\right]_{+},
\omega_{\min},
\omega_{\max}
\Big),
\end{aligned}
$}
\label{eq:entropy_weight}
\end{equation}
where \([z]_{+}=\max(z,0)\), and \(\operatorname{sg}(\cdot)\) denotes
stop-gradient. The weight increases when the modality decision is ambiguous
and decreases after the gate becomes confident. Stop-gradient prevents the
student from changing its entropy merely to manipulate the loss weight.

Combining teacher reliability with student uncertainty, the adaptive gate
distillation loss is defined as
\begin{equation}
\resizebox{0.96\columnwidth}{!}{$\displaystyle
\mathcal{L}_{gate}^{adapt}
=
\sum_l
\frac{
\displaystyle
\sum_{x,y}
R_l(x,y)\omega_l(x,y)
\operatorname{KL}
\left(
Q_l(x,y)\parallel G_l(x,y)
\right)
}{
\displaystyle
\sum_{x,y}
R_l(x,y)\omega_l(x,y)+\epsilon
}.
$}
\label{eq:adaptive_gate_loss}
\end{equation}
Consequently, teacher-reliable locations with ambiguous student decisions
make the largest contribution to gate supervision.

In addition to gate-level supervision, we retain same-modality branch
distillation and opposite-modality feature distillation. Let
\(\mathcal{B}=\{1,2,3,4\}\) denote the four matched backbone stages.
Before feature matching, a \(1\times1\) convolution followed by batch
normalization aligns each student feature with the channel dimension of
the corresponding teacher feature. The two feature-distillation losses
are defined compactly as
\begin{equation}
\begin{aligned}
\mathcal{L}_{branch}
&=
\sum_{l\in\mathcal{B}}
\left[
\mathcal{D}_{\mathrm{cwd}}
\left(\widehat{F}_l^{rgb},\widehat{T}_l^{rgb}\right)
+
\mathcal{D}_{\mathrm{cwd}}
\left(\widehat{F}_l^{ir},\widehat{T}_l^{ir}\right)
\right],\\
\mathcal{L}_{cross}
&=
\sum_{l\in\mathcal{B}}
\left[
\mathcal{D}_{\mathrm{corr}}
\left(\widehat{F}_l^{rgb},\widehat{T}_l^{ir}\right)
+
\mathcal{D}_{\mathrm{corr}}
\left(\widehat{F}_l^{ir},\widehat{T}_l^{rgb}\right)
\right],
\end{aligned}
\end{equation}
where \(\widehat{F}_l^m\) and \(\widehat{T}_l^m\) denote the aligned student
and teacher features, respectively.
\(\mathcal{D}_{\mathrm{cwd}}\) is the channel-wise spatial KL-divergence
loss~\citep{shu2020channel}, and \(\mathcal{D}_{\mathrm{corr}}\) is one minus
the mean channel-wise Pearson correlation after teacher-attention weighting.
Accordingly, \(\mathcal{L}_{branch}\) preserves modality-specific
representations, whereas \(\mathcal{L}_{cross}\) transfers complementary
information from the opposite modality. These losses are applied to the
four backbone stages, while \(\mathcal{L}_{gate}^{adapt}\) supervises the
spatial gates at the three EOGF levels \(P_3\), \(P_4\), and \(P_5\).

The complete training objective is
\begin{equation}
\mathcal{L}_{total}
=
\mathcal{L}_{det}
+
\alpha\lambda(e)
\left(
\mathcal{L}_{branch}
+
\mathcal{L}_{cross}
+
\mathcal{L}_{gate}^{adapt}
\right),
\end{equation}
where \(\mathcal{L}_{det}\) denotes the standard detection loss,
\(\alpha\) is the global distillation weight, and \(\lambda(e)\) controls
the contribution of distillation over the training epochs. Detection-head
and pseudo-fusion distillation are not included in the final objective.

\begin{table*}[t]
\centering
\caption{Split-wise image and annotation statistics before and after DroneVehicle refinement.}
\label{tab:dronevehicle_refinement_stats}
\tabsetup
\begin{adjustbox}{max width=\linewidth}
\begin{tabular}{@{}llrrrrrr@{}}
\toprule
Split & Modality & Orig. Images & Ref. Images & Removed & Orig. Ann. & Ref. Ann. & Delta \\
\midrule
Train & RGB & 17,990 & 17,472 & 518 & 286,793 & 286,444 & -349 \\
Train & IR  & 17,990 & 17,950 & 40  & 316,411 & 316,089 & -322 \\
Val   & RGB & 1,469  & 1,442  & 27  & 22,462  & 22,429  & -33  \\
Val   & IR  & 1,469  & 1,467  & 2   & 24,490  & 24,478  & -12  \\
Test  & RGB & 8,980  & 8,731  & 249 & 143,315 & 143,170 & -145 \\
Test  & IR  & 8,980  & 8,958  & 22  & 159,616 & 159,493 & -123 \\
\midrule
Total & RGB & 28,439 & 27,645 & 794 & 452,570 & 452,043 & -527 \\
Total & IR  & 28,439 & 28,375 & 64  & 500,517 & 500,060 & -457 \\
\bottomrule
\end{tabular}
\end{adjustbox}
\end{table*}

\begin{table*}[t]
\centering
\caption{Category-wise annotation distribution before and after DroneVehicle refinement.}
\label{tab:dronevehicle_category_distribution}
\tabsetup
\begin{adjustbox}{max width=\linewidth}
\begin{tabular}{@{}lrrrrrr@{}}
\toprule
Category & RGB Orig. & RGB Ref. & RGB Delta & IR Orig. & IR Ref. & IR Delta \\
\midrule
car & 389,779 & 379,875 & -9,904 & 428,086 & 417,709 & -10,377 \\
freight car & 13,400 & 11,697 & -1,703 & 17,173 & 14,582 & -2,591 \\
truck & 22,123 & 25,020 & +2,897 & 25,960 & 29,807 & +3,847 \\
bus & 15,333 & 15,341 & +8 & 16,590 & 16,610 & +20 \\
van & 11,935 & 20,110 & +8,175 & 12,708 & 21,352 & +8,644 \\
\bottomrule
\end{tabular}
\end{adjustbox}
\end{table*}

\section{Experiments}
\subsection{Experimental Setup}

We evaluated EGM-Det on three RGB--IR object detection benchmarks with different detection protocols: DroneVehicle~\citep{sun2020drone}, LLVIP~\citep{jia2021llvip}, and VEDAI~\citep{razakarivony2016vehicle}. DroneVehicle was used as the primary benchmark and contains UAV-view image pairs with oriented annotations for five vehicle categories. VEDAI provided an additional aerial benchmark under the oriented bounding box (OBB) protocol, whereas LLVIP was evaluated under the horizontal bounding box (HBB) protocol.

The OBB configuration of EGM-Det was used on DroneVehicle and VEDAI, while the HBB configuration was used on LLVIP. The corresponding RGB and IR teachers adopted the same detector configuration and were trained independently using unimodal inputs. On DroneVehicle and LLVIP, EGM-Det was trained for 132 epochs with an input size of 640 and a batch size of 32. On VEDAI, we followed 10-fold cross-validation and trained each fold for 300 epochs at an input resolution of 1024. The batch size was set to 24, and both unimodal teachers were trained for 300 epochs under the same resolution.

AdamW was used on VEDAI with an initial learning rate of $5\times10^{-4}$ and a weight decay of $5\times10^{-4}$. DroneVehicle and LLVIP followed the default optimizer configuration with an initial learning rate of $1\times10^{-3}$ and the same weight decay. A three-epoch warm-up was used for all datasets. Data augmentation included HSV adjustment, random translation, random scaling, horizontal flipping, and mosaic augmentation. Mosaic augmentation was disabled during the final training stage. All experiments used seed 0 with deterministic initialization, and mixed-precision training was disabled.

We report Precision, Recall, mAP$_{50}$, and mAP$_{50\text{-}95}$ according to the protocol of each dataset. Category-wise AP$_{50}$ is additionally reported on DroneVehicle. All main comparisons and ablation studies on DroneVehicle used the refined annotations. For a fair comparison, all methods in Table~\ref{tab:main_results} used the same data split, category definitions, preprocessing, and OBB evaluation script. Existing methods were retrained using available official implementations or our reproduced implementations under this common protocol. Results obtained with the original annotations were used only to evaluate annotation-protocol sensitivity and were not mixed with the main comparison.

\subsection{DroneVehicle Annotation Refinement}

The original DroneVehicle annotations contain two major forms of noise. Cross-modal inconsistency includes unmatched instances, inconsistent box locations or orientations, and objects annotated in only one modality. Category-level inconsistency mainly occurs among visually similar subclasses, including car, freight car, truck, and van. These errors affect both standard detection supervision and the teacher confidence maps used to construct modality-preference targets.

We refined the annotations by jointly inspecting each RGB image, IR image, and their original labels. Missing or inconsistent instances were corrected only when the paired observations provided reliable evidence. Category labels were changed only when both modalities supported an unambiguous correction. Image entries or object annotations were removed when their modality correspondence or category could not be reliably verified. The task definition, category set, and original train/validation/test split remained unchanged.

The refined samples underwent an additional multi-reviewer quality-control procedure. Reviewers examined cross-modal correspondence, category correctness, and OBB validity using the paired images and the original and refined annotations. Flagged samples were re-examined, unverifiable cases were removed, and all corrected samples were checked again after revision.

Table~\ref{tab:dronevehicle_refinement_stats} summarizes the split-wise changes. The refinement removed 794 RGB entries and 64 IR entries, while the corresponding annotation counts decreased by only 527 and 457. Thus, most annotations were retained while unreliable image entries and labels were corrected or removed. Paired training and evaluation used only valid RGB--IR correspondences with unified labels.

Table~\ref{tab:dronevehicle_category_distribution} reports the category-wise changes. The largest redistributions occurred among car, freight car, truck, and van, whereas the number of bus annotations remained nearly unchanged. These statistics identify category confusion, rather than uniform annotation loss, as a major source of label inconsistency.

\begin{figure*}[!t]
\centering
    \includegraphics[width=0.95\textwidth]{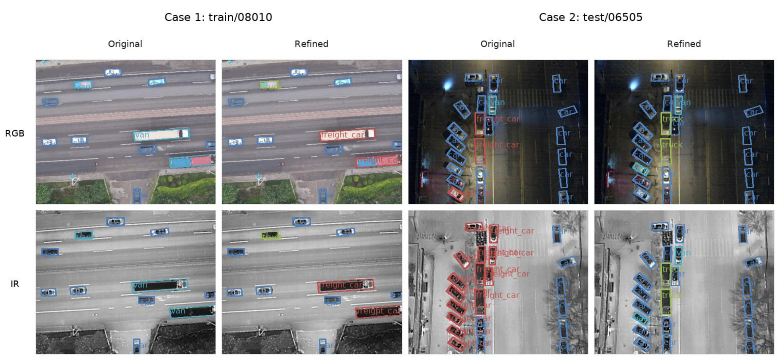}
\caption{
Qualitative examples of DroneVehicle annotation refinement.
Each case compares original and refined annotations on paired RGB and infrared images, highlighting improved cross-modal consistency and category correction.
}
\label{fig:dronevehicle_refinement_examples}
\end{figure*}

Representative corrections are shown in Figure~\ref{fig:dronevehicle_refinement_examples}. The displayed cases illustrate improved cross-modal consistency and corrected category assignments. For reproducibility, the released records contain the split, image identifier, modality, original and refined labels, OBB coordinates, and correction type.

\subsubsection{Sensitivity to the Annotation Protocol}
We trained and evaluated the complete EGM-Det model under both annotation protocols using the same configuration and training schedule. As shown in Table~\ref{tab:annotation_protocol}, the refined protocol increased mAP$_{50}$ from 80.4\% to 85.6\% and mAP$_{50\text{-}95}$ from 67.0\% to 71.4\%. Freight car and van obtained the largest AP$_{50}$ increases, at 9.7 and 11.8 percentage points, respectively. These categories also underwent substantial label correction in Table~\ref{tab:dronevehicle_category_distribution}. Because the protocol changes both training labels and evaluation targets, these differences represent annotation sensitivity rather than an architectural improvement. The refined protocol was therefore used for all subsequent DroneVehicle comparisons and ablations.

\begin{table*}[!htbp]
\centering
\caption{
Sensitivity of EGM-Det to the DroneVehicle annotation protocol. Per-class values are reported
in AP$_{50}$. The original protocol denotes training and evaluation with the original
DroneVehicle annotations, while the refined protocol denotes training and evaluation with
the refined annotations under the same model configuration and training schedule. This
comparison is used to analyze annotation-protocol sensitivity rather than architectural
improvement.
}
\label{tab:annotation_protocol}
\tabsetup
\begin{adjustbox}{max width=\textwidth}
\begin{tabular}{@{}lrrrrrrr@{}}
\toprule
Protocol & Car & Freight Car & Truck & Bus & Van & mAP$_{50}$ & mAP$_{50\text{-}95}$ \\
\midrule
Original annotations & 97.1 & 61.6 & 80.8 & 95.2 & 66.1 & 80.4 & 67.0 \\
Refined annotations  & 98.0 & 71.3 & 85.4 & 95.5 & 77.9 & 85.6 & 71.4 \\
$\Delta$             & +0.9 & +9.7 & +4.6 & +0.3 & +11.8 & +5.2 & +4.4 \\
\bottomrule
\end{tabular}
\end{adjustbox}
\end{table*}

\subsection{Main Results on DroneVehicle}

DroneVehicle is the primary benchmark in this work; therefore, we
provide a detailed category-wise comparison on this dataset. We compare
EGM-Det with two unimodal YOLOv8m baselines~\citep{yolov8_ultralytics}
and representative RGB-IR or multispectral detectors, including
CALNet~\citep{he2023multispectral},
C$^2$Former~\citep{yuan2024c2former},
E2E-MFD~\citep{zhang2024e2emfd},
ICAFusion~\citep{shen2024icafusion},
COMO~\citep{liu2026cross},
and LCAFNet~\citep{wu2026lightweight}.
The compared methods are arranged chronologically by publication year
in Table~\ref{tab:main_results}. The Year/Venue column indicates the
original publication year and venue of each method, whereas all
performance values are obtained by retraining and evaluating the
methods under the same refined DroneVehicle protocol.

In addition to the overall mAP$_{50}$ and
mAP$_{50\text{-}95}$, we report AP$_{50}$ for each vehicle category,
including car, freight car, truck, bus, and van. This fine-grained
evaluation allows us to analyze not only overall detection performance,
but also the behavior of different methods on visually similar and
scale-varying vehicle categories.

\begin{table*}[!htbp]
\centering
\caption{
Main results on the refined DroneVehicle benchmark, arranged
chronologically by publication year. Per-class results are reported
in AP$_{50}$. The Year/Venue column indicates the original publication
source of each method, while all methods are retrained and evaluated
under the same refined DroneVehicle protocol.
}
\label{tab:main_results}
\tabsetup
\begin{adjustbox}{max width=\textwidth}
\begin{tabular}{@{}llrrrrrrr@{}}
\toprule
Method & Year/Venue & Car & Freight Car & Truck & Bus & Van
& mAP$_{50}$ & mAP$_{50\text{-}95}$ \\
\midrule

YOLOv8m (RGB)
& 2023/Ultralytics
& 93.3 & 45.2 & 63.3 & 88.0 & 70.3 & 72.0 & 54.1 \\

YOLOv8m (IR)
& 2023/Ultralytics
& 96.7 & 54.6 & 72.2 & 90.4 & 66.5 & 76.1 & 60.8 \\

CALNet
& 2023/ACM MM
& 97.2 & 70.5 & 76.8 & 96.7 & 69.0 & 82.0 & 57.4 \\

C$^2$Former
& 2024/TGRS
& 89.6 & 61.5 & 70.9 & 88.1 & 74.6 & 76.9 & 45.1 \\

E2E-MFD
& 2024/NeurIPS
& 88.6 & 48.2 & 60.1 & 77.6 & 65.1 & 67.9 & 41.8 \\

ICAFusion
& 2024/PR
& 97.9 & 70.6 & 84.0 & 96.6 & 82.3
& \textbf{86.2} & 62.4 \\

LCAFNet
& 2026/PR
& 97.5 & 70.2 & 83.2 & 95.2 & 82.5
& 85.7 & 59.9 \\

COMO
& 2026/INFORM FUSION
& 97.4 & 70.5 & 84.2 & 95.8 & 81.7
& \underline{85.9} & \underline{65.5} \\

\hline
EGM-Det
& Ours
& 98.0 & 71.3 & 85.4 & 95.5 & 77.9
& 85.6 & \textbf{71.4} \\

\bottomrule
\end{tabular}
\end{adjustbox}
\end{table*}

As shown in Table~\ref{tab:main_results}, under the unified refined
DroneVehicle protocol, EGM-Det achieves the best
mAP$_{50\text{-}95}$ of 71.4\%, outperforming the strongest competing
method, COMO, by 5.9 percentage points. This improvement indicates
that EGM-Det provides stronger localization robustness under stricter
IoU thresholds, rather than only improving loose-threshold detections.

At the mAP$_{50}$ level, EGM-Det obtains 85.6\%, which is competitive
with ICAFusion (86.2\%), COMO (85.9\%), and LCAFNet (85.7\%).
In terms of category-wise AP$_{50}$, EGM-Det achieves the best
performance on car, freight car, and truck. For bus and van,
CALNet and LCAFNet achieve the highest AP$_{50}$, respectively.
These results indicate that EGM-Det provides a clear advantage under
stricter localization criteria, while further improvement is still
possible for category-level discrimination on certain vehicle
subclasses.

The comparison with unimodal YOLOv8m baselines further confirms the necessity of
RGB-IR fusion. The IR-only model outperforms the RGB-only model in overall performance,
especially on freight car and truck, but both unimodal detectors are clearly inferior to
strong multimodal approaches. Compared with existing multimodal detectors, EGM-Det
explicitly supervises modality preference during feature fusion through dual-teacher
guidance and student-gate-entropy adaptive weighting. The results support our central
motivation: reliable RGB-IR detection requires not only combining multimodal features,
but also learning where and how each modality should contribute.

\subsection{Additional Results on LLVIP and VEDAI}

To further evaluate the applicability of EGM-Det across different
RGB-IR detection protocols, we conduct additional evaluations on LLVIP
and VEDAI. Unlike DroneVehicle, which serves as the primary benchmark
and is evaluated with detailed category-wise results, these two datasets
are used as complementary benchmarks. LLVIP follows the horizontal
bounding box (HBB) setting, while VEDAI adopts the aerial oriented
bounding box (OBB) setting. Table~\ref{tab:generalization_llvip_vedai}
summarizes representative results using mAP$_{50\text{-}95}$ and model
parameters.

For LLVIP and VEDAI, we compare EGM-Det with representative
unimodal detectors, including YOLOv5m~\citep{yolov5_ultralytics},
YOLOv8m~\citep{yolov8_ultralytics}, and
DINO~\citep{zhang2022dino}. The multimodal comparisons include
CFT~\citep{fang2021cross},
SuperYOLO~\citep{zhang2022superyolo},
ICAFusion~\citep{shen2024icafusion},
CMADet~\citep{song2024misaligned},
EFAF~\citep{peng2025efaf},
FQDNet-s~\citep{meng2025fqdnet},
LMDENet~\citep{weng2026lmdenet},
ISDG-Net~\citep{gao2026isdgnet},
and MCF-YOLO~\citep{yang2026mcfyolo}.

\begin{table}[!htbp]
\centering
\caption{
Additional results on LLVIP and VEDAI, arranged chronologically by
publication year. A dash indicates an unavailable result under the
corresponding evaluation setting. The best and second-best results
in each dataset column are highlighted in bold and underlined,
respectively.
}
\label{tab:generalization_llvip_vedai}
\tabsetup

\begin{adjustbox}{max width=\columnwidth}
\begin{tabular}{@{}lllrrr@{}}
\toprule
Method & Modality & 
LLVIP & VEDAI \\
\midrule




DINO

& RGB & 53.8 & 44.9 \\

YOLOv8m

& RGB & 49.2 & 37.8 \\
\hline

DINO

& IR & 62.9 & 41.7 \\

YOLOv8m

& IR & 62.8 & 35.3 \\
\hline
SuperYOLO

& RGB+IR & 58.1 & \underline{48.3} \\

ICAFusion

& RGB+IR & 47.9 & 45.2 \\

CMADet

& RGB+IR
& 62.9
& -- \\

EFAF

& RGB+IR
& 63.4
& 47.7 \\

FQDNet-s

& RGB+IR
& \underline{64.1}
& 47.7 \\

LMDENet

& RGB+IR
& 59.2
& -- \\

ISDG-Net

& RGB+IR
& 63.6
& -- \\

MCF-YOLO

& RGB+IR
& --
& 45.0 \\

Ours (EGM-Det)
& RGB+IR
& \textbf{64.2}
& \textbf{60.3} \\

\bottomrule
\end{tabular}
\end{adjustbox}
\end{table}

As shown in Table~\ref{tab:generalization_llvip_vedai},
EGM-Det achieves 64.2\% mAP$_{50\text{-}95}$ on LLVIP, yielding
the highest result among the compared methods. Although the margin
over the strongest recent competitor is relatively small, EGM-Det
consistently performs favorably against both recent RGB-IR detectors
and representative multimodal baselines. This result demonstrates
the effectiveness of the proposed adaptive fusion and
modality-preference supervision under the LLVIP HBB setting.

On VEDAI, EGM-Det obtains 60.3\%
mAP$_{50\text{-}95}$, achieving the best performance among the
methods with available results under the adopted OBB evaluation
protocol. It exceeds the second-highest reported result by 12.0 percentage
points and maintains a clear advantage over both representative
baselines and recently published RGB-IR detectors.

Overall, the LLVIP and VEDAI results complement the primary
DroneVehicle evaluation. EGM-Det achieves competitive performance
under the LLVIP HBB setting and obtains the best result among the
methods with available VEDAI results. These evaluations demonstrate
that the proposed modality-preference supervision and
student-gate-entropy adaptive weighting can be applied across
different RGB-IR detection settings.

\begin{figure*}[!t]
    \centering
    \includegraphics[width=\textwidth]{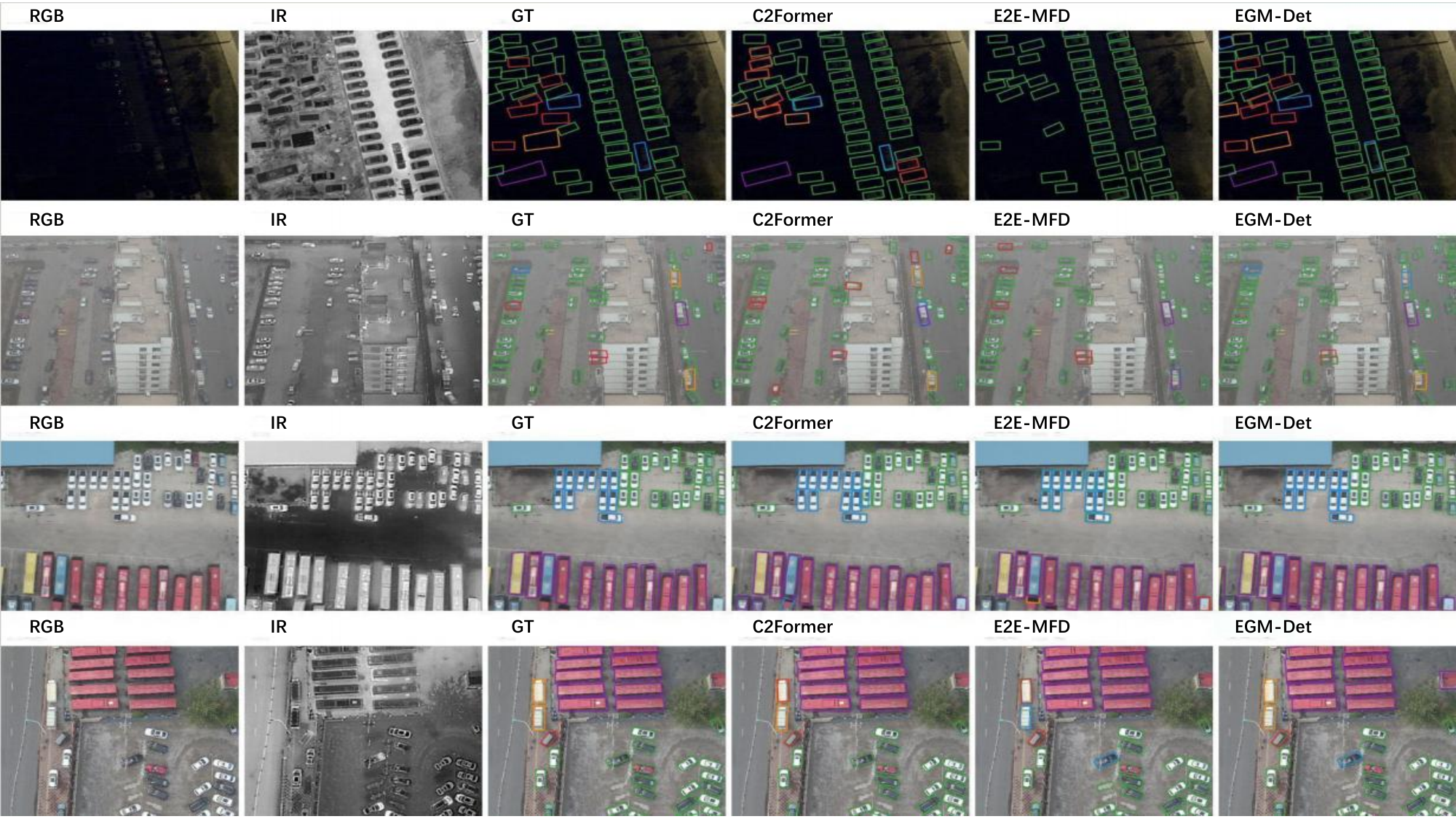}
    \vspace{-2mm}
    \caption{
    Qualitative detection comparison on representative DroneVehicle test samples. From left to right: RGB image, infrared image, ground truth, C2Former, E2E-MFD, and EGM-Det. For clarity, only oriented bounding boxes are shown without per-box text. EGM-Det yields more complete detections in low-light and dense scenes while reducing missed objects and category confusion. Colors denote categories: \legendbox{green} car, \legendbox{orange} freight car, \legendbox{blue} truck, \legendbox{purple} bus, and \legendbox{red} van.}
    \label{fig:qual_detection}
\end{figure*}

\begin{table*}[!htbp]
\centering
\caption{
Ablation study of EGM-Det on the refined DroneVehicle test set. Per-class values are reported
in AP$_{50}$.
}
\label{tab:egmdet_ablation}
\tabsetup
\begin{adjustbox}{max width=\textwidth}
\begin{tabular}{@{}lrrrrrrr@{}}
\toprule
Method & Car & Freight Car & Truck & Bus & Van & mAP$_{50}$ & mAP$_{50\text{-}95}$ \\
\midrule
Ours (EGM-Det) & 98.0 & 71.3 & 85.4 & 95.5 & 77.9 & 85.6 & 71.4 \\
Ours w/o student entropy weighting & 97.9 & 70.6 & 84.3 & 95.2 & 75.4 & 84.7 & 70.6 \\
Ours w/o dual-teacher distillation & 97.6 & 70.3 & 83.9 & 95.1 & 75.1 & 84.4 & 70.4 \\
Ours w/o entropy-guided fusion and distillation & 97.4 & 67.9 & 81.3 & 94.8 & 74.3 & 83.1 & 68.4 \\
\bottomrule
\end{tabular}
\end{adjustbox}
\end{table*}

\subsection{Ablation Studies}

To further analyze the contribution of each component in EGM-Det, we conduct ablation
studies on the refined DroneVehicle test set. The ablation variants are designed to examine
the effects of the proposed fusion architecture, dual-teacher distillation, and
student-entropy-based gate reweighting. Specifically, we evaluate three variants:
(1) removing the entropy-guided fusion architecture and distillation, where the student
is reduced to a naive feature-summation multimodal baseline; (2) retaining the
entropy-guided fusion student but removing dual-teacher distillation; and (3) keeping
dual-teacher distillation while disabling student-entropy-based gate reweighting. This
setting allows us to examine the contribution of the fusion architecture, teacher-derived
modality-preference supervision, and entropy-adaptive gate weighting.

\begin{figure}
    \centering
    \includegraphics[width=\columnwidth]{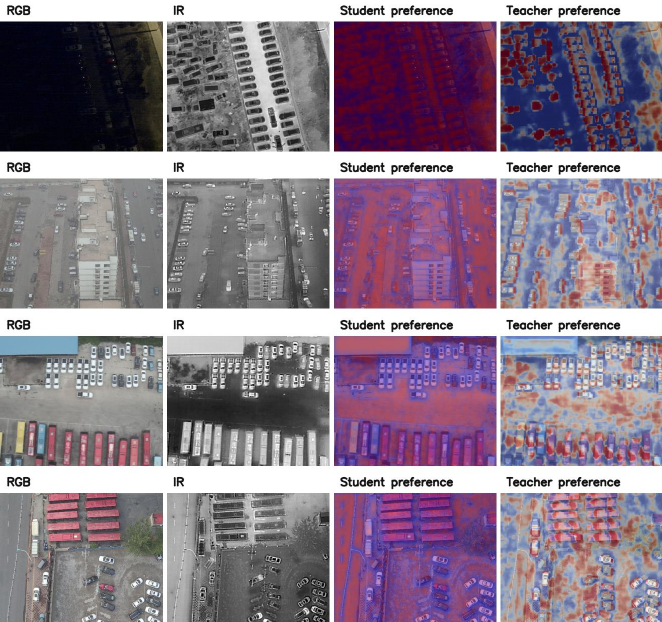}
    \vspace{-2mm}
    \caption{Visualization of modality preference in EGM-Det. Blue/red indicate RGB-/IR-preferred regions; student and teacher maps show spatially adaptive modality reliability across representative DroneVehicle scenes.}
    \label{fig:qual_gate}
\end{figure}

As shown in Table~\ref{tab:egmdet_ablation}, the most reduced baseline, which removes
entropy-guided fusion and distillation, suffers the largest degradation, reducing
mAP$_{50\text{-}95}$ from 71.4\% to 68.4\% and mAP$_{50}$ from 85.6\% to 83.1\%.
This confirms that naive RGB-IR feature summation is insufficient for robust UAV-view
multimodal detection. In particular, this baseline shows clear degradation on visually
ambiguous vehicle categories such as freight car, truck, and van, suggesting that local
cross-modal discrepancy, alignment confidence, and spatial modality preference should be
explicitly modeled rather than treated as uniform feature fusion.

Removing dual-teacher distillation while retaining the entropy-guided student decreases
mAP$_{50\text{-}95}$ from 71.4\% to 70.4\% and mAP$_{50}$ from 85.6\% to 84.4\%.
This result indicates that the proposed student architecture is already strong under
standard detection supervision, while teacher-derived modality preference still provides
complementary guidance. The gain from dual-teacher distillation is moderate rather than
dominant, which is consistent with the selective design of EGM-Det: the teachers are not used
to force the student to mimic all predictions, but to regularize modality-specific
representations, cross-modal complementarity, and gate-level selection behavior.

Finally, disabling student-entropy-based weighting reduces mAP$_{50\text{-}95}$ from
71.4\% to 70.6\% and mAP$_{50}$ from 85.6\% to 84.7\%. Although the overall gain is
moderate, entropy weighting improves several category-level results. For example, van
AP$_{50}$ increases from 75.4\% to 77.9\%, and truck AP$_{50}$ increases from 84.3\% to
85.4\% when entropy weighting is enabled. This suggests that student entropy weighting
acts as a fine-grained stabilizer for gate distillation. Instead of uniformly enforcing
teacher-derived gate objects over all spatial locations, EGM-Det adaptively strengthens
supervision in uncertain regions and relaxes it when the student has already formed
confident modality decisions.

Overall, the ablation results support the effectiveness of the proposed design. The
entropy-guided fusion architecture provides the main performance gain, dual-teacher
distillation further improves the student by supervising modality preference, and
student-entropy weighting refines gate supervision in an uncertainty-aware manner. These
findings support the central motivation of EGM-Det: effective RGB-IR detection requires not
only cross-modal feature fusion, but also explicit and selective learning of where and how each
modality should contribute.

\begin{figure}
    \centering
    \includegraphics[width=\columnwidth]{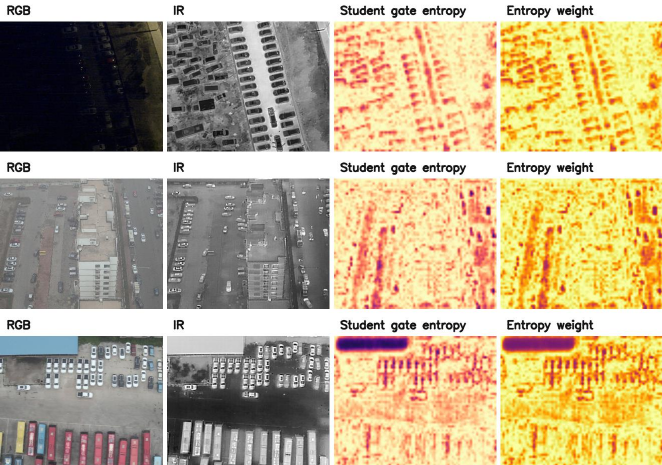}
    \vspace{-2mm}
    \caption{Visualization of student gate entropy and entropy-induced weighting. High-entropy regions indicate uncertain modality selection, and the corresponding weight map highlights locations where cross-modal reasoning is most needed.}
    \label{fig:qual_entropy}
\end{figure}

\subsection{Qualitative Analysis}
\label{sec:qualitative_analysis}

To further examine the behavior of EGM-Det beyond quantitative metrics, we provide qualitative results on representative DroneVehicle test samples, covering low-light scenes, dense parking areas, mixed vehicle categories, and modality-degraded conditions. As shown in Figure~\ref{fig:qual_detection}, the compared methods exhibit different failure patterns under these challenging scenarios. C2Former and E2E-MFD can localize many salient vehicles, but they still suffer from missed detections in low-contrast regions and occasional category confusion among visually similar classes, such as trucks, vans, and freight cars. In contrast, EGM-Det produces more complete oriented detections across both sparse and dense regions, especially in scenes where RGB appearance is weak but infrared responses remain informative. This suggests that EGM-Det does not simply increase the number of predictions, but improves the reliability of cross-modal evidence aggregation for small and densely distributed oriented objects.

We further visualize the learned modality preference of EGM-Det in Figure~\ref{fig:qual_gate}. Blue regions indicate stronger RGB preference, whereas red regions indicate stronger infrared preference. The learned student gate shows clear spatial adaptivity: regions with reliable thermal contrast tend to receive stronger infrared emphasis, while regions where visible texture remains informative preserve RGB contributions. This behavior is particularly important for nighttime DroneVehicle imagery, where RGB images may contain rich structural cues in illuminated regions but become unreliable in shadows or low-exposure areas. The teacher preference maps provide an additional reference for cross-modal supervision. The consistency between teacher preference and student gate responses indicates that EGM-Det successfully transfers modality-specific reliability from the dual teachers to the compact student model.

Figure~\ref{fig:qual_entropy} visualizes the student gate entropy and the corresponding entropy weight. High-entropy regions mainly appear near object boundaries, dense vehicle clusters, low-contrast areas, and regions where RGB and infrared cues are inconsistent. These are precisely the locations where modality selection is more ambiguous and where naive fusion is more likely to introduce noisy or biased representations. The entropy weight map amplifies such uncertain regions, encouraging the student to pay more attention to difficult cross-modal fusion cases during distillation. This provides an intuitive explanation for the effectiveness of the student entropy weighting strategy: rather than treating all spatial locations equally, EGM-Det emphasizes regions where modality reliability is harder to determine.

Overall, these qualitative observations are consistent with the quantitative comparisons and ablation studies. The detection visualizations show that EGM-Det improves the completeness and stability of oriented vehicle detection, while the gate preference and entropy maps reveal why the improvement occurs: EGM-Det learns spatially adaptive modality selection and further regularizes uncertain fusion regions through entropy-aware distillation. These results support the central design of EGM-Det, namely combining entropy-guided fusion with dual-teacher distillation to obtain a more robust multispectral detector.

\section{Discussion}
The results indicate that EGM-Det benefits primarily from reliability-aware modality selection rather than indiscriminate feature aggregation. On the refined DroneVehicle protocol, EGM-Det achieves the highest mAP$_{50\text{-}95}$ among the compared methods while maintaining competitive mAP$_{50}$. This result suggests that the proposed framework is particularly effective under stricter localization criteria, which are important for small, dense, and arbitrarily oriented objects in UAV-view scenes.

The ablation results further reveal the complementary roles of EntropyOffsetGateFusion (EOGF) and selective multimodal distillation. Compared with naive feature summation, EOGF accounts for most of the improvement by introducing entropy-guided alignment and spatially adaptive fusion. Dual-teacher supervision provides an additional gain by converting relative teacher confidence into explicit modality-preference targets. Student-gate-entropy weighting further directs this supervision toward locations with uncertain modality decisions. The visualization results support this interpretation: infrared information is emphasized in thermally salient or poorly illuminated regions, whereas RGB information remains dominant where visible structures are more reliable.

The comparison between the original and refined DroneVehicle protocols should be interpreted as annotation sensitivity rather than an architectural improvement. The larger performance changes for freight car and van correspond to categories receiving substantial annotation correction. This observation indicates that cross-modal inconsistency and category-level label noise can affect both detection supervision and teacher-derived modality-preference targets. Separately reporting results under the refined protocol is therefore necessary for reproducible and fair comparison.

Several limitations should be considered when interpreting the current results and their broader applicability. First, teacher confidence is only a proxy for modality reliability and may become biased when the teachers are poorly calibrated or applied to out-of-distribution scenes~\citep{guo2017calibration}. Second, the dual-teacher design increases training cost, although only the dual-stream detector is retained during inference. Finally, gate entropy represents uncertainty in modality selection but does not fully describe uncertainty caused by sensor noise, domain shift, or annotation ambiguity~\citep{kendall2017uncertainties}. Future work will investigate calibrated or teacher-free preference learning and extend the framework to temporal and cross-domain RGB--IR detection.

\section{Conclusions}
This paper presented EGM-Det, an entropy-guided adaptive fusion framework for UAV-view RGB--IR object detection. Its EOGF module combines input-derived entropy priors, local feature alignment, and spatial-channel gating to aggregate complementary multimodal information. During training, two unimodal teachers convert their relative confidence into modality-preference targets, while student-gate entropy adaptively adjusts the strength of gate supervision.
EGM-Det achieves 85.6\% mAP$_{50}$ and 71.4\% mAP$_{50\text{-}95}$ on the refined DroneVehicle protocol. Evaluations on LLVIP and VEDAI further demonstrate its applicability to both horizontal and oriented object detection. The ablation results confirm that entropy-guided fusion provides the primary improvement, while selective gate supervision further refines local modality selection. These findings show that explicitly modeling spatially varying modality reliability is beneficial for UAV-view RGB--IR detection. Nevertheless, the framework may remain sensitive to teacher calibration and paired-data quality, while dual teachers introduce additional training cost. Future work will explore more efficient preference learning and broader temporal and cross-domain evaluation.

\section*{CRediT authorship contribution statement}

Conceptualization, C.F., D.Y. and H.Z.; methodology, C.F., D.Y., G.W. and H.Z.;
software, C.F. and D.Y.; validation, C.F., D.Y., G.W., X.Y. and Y.J.; data
curation, C.F., D.Y. and G.W.; writing -- original draft, C.F. and D.Y.; writing
-- review and editing, all authors; visualization, C.F., D.Y. and G.W.;
supervision, J.Y. and H.Z.; funding acquisition, H.Z. All authors have read
and agreed to the published version of the manuscript.

\section*{Funding}

This work was supported by the National Natural Science Foundation of China
[grant number 62401471] and the 2024 Gusu Innovation and Entrepreneurship
Leading Talents Program (Young Innovative Leading Talents)
[grant number ZXL2024333].

\section*{Data availability}

The DroneVehicle, VEDAI, and LLVIP datasets are available from their original
providers. The EGM-Det code and refined DroneVehicle annotations are available
in the \href{https://github.com/NowhereDuk/SEMD-Selective-Entropy-guided-Multimodal-Distillation/tree/clean-main}{public project repository}.

\section*{Declaration of competing interest}

The authors declare no conflicts of interest.

\section*{Declaration of generative AI and AI-assisted technologies in the writing process}

The authors used ChatGPT (OpenAI) only to polish the language during manuscript
preparation. The authors reviewed and edited the manuscript and take full
responsibility for its content.

\bibliographystyle{cas-model2-names}
\balance
\bibliography{references}

\end{document}